\documentclass[10pt,twocolumn,letterpaper]{article}
\usepackage{cvpr}

\usepackage{eccvabbrv}

\usepackage{float}
\usepackage{wrapfig}
\usepackage{graphicx}
\usepackage{booktabs}
\usepackage{multirow}
\usepackage{booktabs}
\usepackage{multirow}
\usepackage{colortbl}
\usepackage{xcolor}
\usepackage{pifont}
\usepackage{xspace}
\usepackage{graphicx} 
\usepackage{adjustbox}

\newcommand{\method}{WorldFlow3D}

\usepackage[accsupp]{axessibility}  

\usepackage{hyperref}

\usepackage{orcidlink}

\title{\hspace{-5pt}WorldFlow3D: Flowing Through 3D Distributions\\ for Unbounded World Generation} 

\author{
    Amogh Joshi\textsuperscript{\rm 1}$^*$,\;
    Julian Ost\textsuperscript{\rm 1}$^*$,\;
    Felix Heide\textsuperscript{\rm 1, 2}
    \\ \\
    \textsuperscript{\rm 1}Princeton University \quad \textsuperscript{\rm 2}Torc Robotics 
}

\begin{document}
\twocolumn[{%
\renewcommand\twocolumn[1][]{#1}%
\vspace{-8mm}
\maketitle
\begin{center}
\centering
\includegraphics[width=\linewidth]{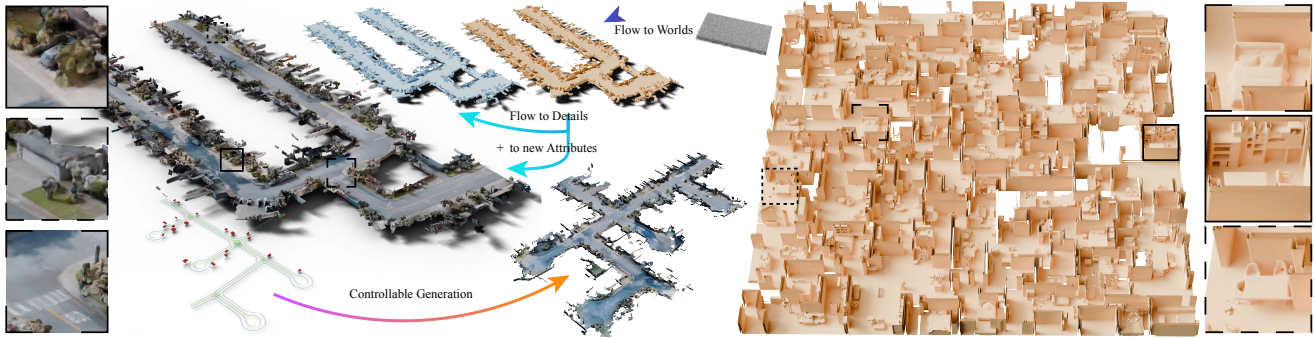}
\captionof{figure}{\textbf{\method} is a novel method for the generation of unbounded 3D worlds. We show the capabilities of \method~for the generation of large-scale outdoor and indoor scenes, with insets showing learned distributions of fine geometric detail and realistic texture.}
\label{fig:teaser}
\end{center}
}]
\renewcommand{\thefootnote}{\fnsymbol{footnote}}
\footnotetext[1]{Equal contribution.}
\renewcommand{\thefootnote}{\arabic{footnote}}
\begin{abstract}
Unbounded 3D world generation is emerging as a foundational task for scene modeling in computer vision, graphics, and robotics. In this work, we present~\method, a novel method capable of  generating unbounded 3D worlds.  Building upon a foundational property of flow matching -- namely, defining a path of transport between two data distributions -- we model 3D generation more generally as a problem of flowing through 3D data distributions, not limited to conditional denoising. We find that our latent-free flow approach generates causal and accurate 3D structure, and can use this as an intermediate distribution to guide the generation of more complex structure and high-quality texture -- all while converging more rapidly than existing methods.  We enable controllability over generated scenes with vectorized scene layout conditions for geometric structure control and visual texture control through scene attributes. We confirm the effectiveness of~\method~on both real outdoor driving scenes and synthetic indoor scenes, validating cross-domain generalizability and high-quality generation on real data distributions. We confirm favorable scene generation fidelity over approaches in all tested settings for unbounded scene generation. For more, see~\url{https://light.princeton.edu/worldflow3d}.
\end{abstract}
\section{Introduction}
Developing spatially intelligent systems in large-scale environments has long been a central pursuit in computer vision and robotics. 
A core display of intelligence in this context is the ability to synthesize and reason over realistic 3D models of the real world.
This implicitly demonstrates coherent world understanding and processing of spatial relationships centered around visual and geometric causality. 
A growing body of recent work has enabled high-quality \emph{3D reconstructions} from real-world scene captures~\cite{schonberger2016structure,agarwal2011building}. 
Recent learned neural scene representations are capable of producing both implicit~\cite{mildenhall2021nerf,barron2023zipnerf} and explicit~\cite{kerbl3Dgaussians,Huang20242DGS} 3D models from images. 
Scene modeling by reconstruction, however, is fundamentally constrained by a reliance on real data -- naturally translating into a need for purely generative approaches for producing unlimited data.

Modern \emph{3D generation} approaches~\cite{xiang2024trellis, hunyuan3d22025tencent} have shown great success in object-level generation, with high fidelity in both structure and visual texture. 
However, modeling large-scale, realistic 3D \textit{scenes} requires a distinct level of 3D scene understanding, consisting of objects within a broader spatial domain and environmental context.
Procedural modeling methods are capable of producing theoretically unbounded scenes~\cite{infinigen2024indoors,infinitegen, lin2023infinicity, 3dgpt}, but their hand-crafted rule-based approach comes at the cost of not only photorealism in texture but also realism in structure.
Real-world environments, in contrast, exhibit vast diversity in scale, structure, and appearance. 
Some works have shown the ability to model large synthetic environments~\cite{meng2024lt3sdlatenttrees3d,worldgrow2025, wu2024blockfusion}, but real-world, open-world scenes are more complex.
Open-world outdoor scenes, such as driving scenes~\cite{sun2020wod}, are fundamentally structurally sparse -- preventing existing unbounded approaches from translating to such environments.
More recent scene-focused 3D generation approaches use hierarchical 3D latent diffusion~\cite{wu2024blockfusion,meng2024lt3sdlatenttrees3d,zyrianov2025lidardm, lee2025nuiscene}, but are either constrained to a specific data distribution~\cite{meng2024lt3sdlatenttrees3d}, prohibiting generalizability, or are fixed in spatial extent~\cite{ren2024xcube}. Therefore, as summarized in Table~\ref{tab:method_comparison}, a method capable of producing unbounded scenes with high-fidelity geometry and texture, \textit{and} allowing full controllability across domains, remains an open challenge.

We introduce~\method, a novel approach for generating unbounded 3D worlds with full controllability.
~\method~is built on a foundational property of flow matching~\cite{chen2018neural,lipman2022flow} -- defining a path of transport between \textit{any} two data distributions.
Building upon this, we reformulate 3D generation not as a problem of progressive hierarchical conditional denoising but as flowing through sequential 3D data distributions. As such, ~\method~directly produces volumetric scene representations successively from noise, through coarse structure, and into detailed, causal geometry and high-fidelity texture -- all modeled as transport through data distributions. Our approach allows for latent-free, purely volumetric generative models, breaking from the standard autoencode $\to$ generate paradigm of existing methods~\cite{meng2024lt3sdlatenttrees3d, ren2024xcube, xiang2024trellis}. 


We validate~\method~for unbounded scene generations across real, open-world outdoor driving scenes~\cite{sun2020wod} and synthetic indoor rooms~\cite{3dfront} -- confirming that our method obtains high quality across multiple distinct data distributions.
We allow for explicit 3D controllability through vectorized scene layouts for structure and scene attributes for texture.
Our flow matching formulation also enables rapid, latent-free training convergence even on complex 3D data distributions, and efficient inference for generating large-scale worlds.
We introduce an extension to existing schedulers by aligning predicted flow fields across smaller chunks at inference time, unlocking truly unbounded scene generation (limited only by compute) without visible border artifacts.
We measure major improvements upon all existing tested methods, across multiple data domains. Additionally, visual analysis and a blind user study confirms the significance of our results qualitatively.

We summarize our contributions as follows:
\begin{itemize}
    \item We introduce a novel 3D world generation method that formulates 3D generation as flow matching across 3D data distributions.
    \item Our proposed method allows for latent-free generation of scenes with (a) unbounded spatial extent, (b) high-quality geometric structure and realistic visual texture, (c) full controllability over scene layout and visual attributes.
    \item We validate our method with large-scale generations across domains, confirming favorable 3D geometric and texture fidelity in all experiments.
\end{itemize}

\section{Related Work}
\paragraph{3D Object Generation and Procedural Scene Generation.}
Recent advances in 3D object generation have demonstrated remarkable capabilities in synthesizing high-quality textured 3D assets. Object generation methods~\cite{xiang2024trellis, hunyuan3d22025tencent}, and more generally diffusion-based~\cite{hunyuan3d22025tencent,topia,hyperdiffusion,3dgen,rodin,lion,ssdnerf} and transformer-based~\cite{meshgpt,meshanything,meshlrm} methods, have recently shown that generative priors learned from large-scale datasets allow for producing realistic object-level geometry and appearance with explicit control. These methods establish a foundation for generative 3D modeling, but remain limited to isolated objects or bounded spatial contexts. As such, they cannot directly scale to complex scene-level synthesis involving multiple entities, spatial layouts, and environmental context.

\paragraph{Scene Generation.}
Early examples of simulated worlds have been crafted as manual assets~\cite{dosovitskiy2017carla}, thus enabling large-scale experimentation, composition with dynamic actors, and replayable 
evaluation of perception models.
\newcommand{\rbad}{\cellcolor{red!15}{\color{red}\ding{55}}}
\newcommand{\rgood}{\cellcolor{green!15}{\color{green!60!black}\ding{51}}}
\newcolumntype{C}{>{\centering\arraybackslash}p{1.1em}}

\newcommand{\rothead}[1]{%
  \rotatebox{60}{\strut #1}%
}

\begin{wraptable}{r}{0.57\linewidth}
\centering
\scriptsize
\setlength{\tabcolsep}{2pt}
\renewcommand{\arraystretch}{1.0}
\begin{tabular}{>{\raggedright\arraybackslash}p{6em} *{8}{C}}
\toprule
& \rothead{XCube~\cite{ren2024xcube}}
& \rothead{BlockFusion~\cite{wu2024blockfusion}}
& \rothead{LidarDM~\cite{zyrianov2025lidardm}}
& \rothead{LT3SD~\cite{meng2024lt3sdlatenttrees3d}}
& \rothead{InfiniCube~\cite{lu2024infinicube}}
& \rothead{WorldGrow~\cite{worldgrow2025}}
& \rothead{XScene~\cite{yang2025xscene}}
& \rothead{\textbf{Ours}} \\ 
\midrule
Unbounded    & \rbad  & \rgood & \rbad  & \rgood & \rgood & \rgood & \rgood & \rgood \\
Controllable & \rbad  & \rbad  & \rbad  & \rbad  & \rgood & \rbad  & \rbad  & \rgood \\
Cross-Domain & \rgood & \rgood & \rbad  & \rbad  & \rbad  & \rbad  & \rbad  & \rgood \\
Texture      & \rbad  & \rbad  & \rbad  & \rbad  & \rgood & \rgood & \rgood & \rgood \\
\bottomrule
\end{tabular}
\caption{\textbf{Summary of recent 3D scene generation methods.} Ours is the only approach satisfying all desirable criteria.
}
\label{tab:method_comparison}
\end{wraptable}
\noindent However, hand-crafted 3D design is prohibitively expensive and time-consuming.
Procedural modeling approaches~\cite{infinigen2024indoors,infinitegen,lin2023infinicity,3dgpt} have been proposed to resolve this bottleneck, but at the cost of photorealism and variability, both critical aspects of simulation efficacy. 
Some approaches have built on existing 3D object-centric approaches and integrate multiple components together to create pipelines for block-wise 3D world construction~\cite{engstler2025syncity, chen2025trellisworld}, yet these are limited by individual component cohesiveness and broadly lack real 3D awareness.

\paragraph{Hierarchical Latent Scene Generation.}
Building upon object-level generative models, recent works~\cite{ren2024xcube,ren2024scube,text2room,lu2024infinicube,wu2024blockfusion,meng2024lt3sdlatenttrees3d,lsd3d} have extended 3D generation to scene-level synthesis for both indoor and outdoor environments. XCube~\cite{ren2024xcube} sets a benchmark for 3D generation quality via a multi-resolution sparse voxel hierarchy, while SCube~\cite{ren2024scube} and InfiniCube~\cite{lu2024infinicube} introduce controllability and texture modeling. Nevertheless, they remain limited in spatial extent or geometric fidelity.
BlockFusion~\cite{wu2024blockfusion} represents scenes as latent tri-planes and performs spatial extrapolation for larger-scale outpainting, while LT3SD~\cite{3dfront} introduces a latent tree-structured representation for patch-wise geometry synthesis over expansive environment.
However, LT3SD is explicitly confined to dense indoor scenes and neither are capable of appearance modeling.
In the outdoor domain, LidarDM~\cite{zyrianov2025lidardm} generates LiDAR via underlying 3D scene modeling, yet remains limited in fidelity and scope.
WoVoGen~\cite{wovogen} and XScene~\cite{yang2025xscene} explore joint voxel-based occupancy and image generation, but struggle with geometry-texture alignment at scale.
The very recent LSD-3D~\cite{lsd3d} produces high-quality scene textures, but depends on the above methods to supply proxy geometry.

\vspace{1mm}
\noindent Overall, prior works are limited in some combination of fidelity, texture synthesis, spatial extent, or controllability. We propose a novel controllable and unbounded 3D generation method that generalizes across domains, situated among recent works in Table~\ref{tab:method_comparison}.

\section{WorldFlow3D}
\label{sec:method}

In this section, we introduce~\method, a novel formulation of the hierarchical 3D generation problem via flow matching.
We formulate our method as transport between hierarchical distributions via flow models in~\cref{ssec:flows}. We propose to \emph{directly generate} volumetric distributions of scene surfaces in~\cref{ssec:gen}, where we revisit autoencoder-free generation departing from latent diffusion models.
Finally, we describe how~\method~can be used to perform controllable (\ref{ssec:control}) and unbounded (\ref{ssec:unbounded}) 3D world generation through conditional and inference-time flow field alignment across smaller chunks.

\begin{figure*}[t!]
    \centering
    \includegraphics[width=1.0\linewidth]{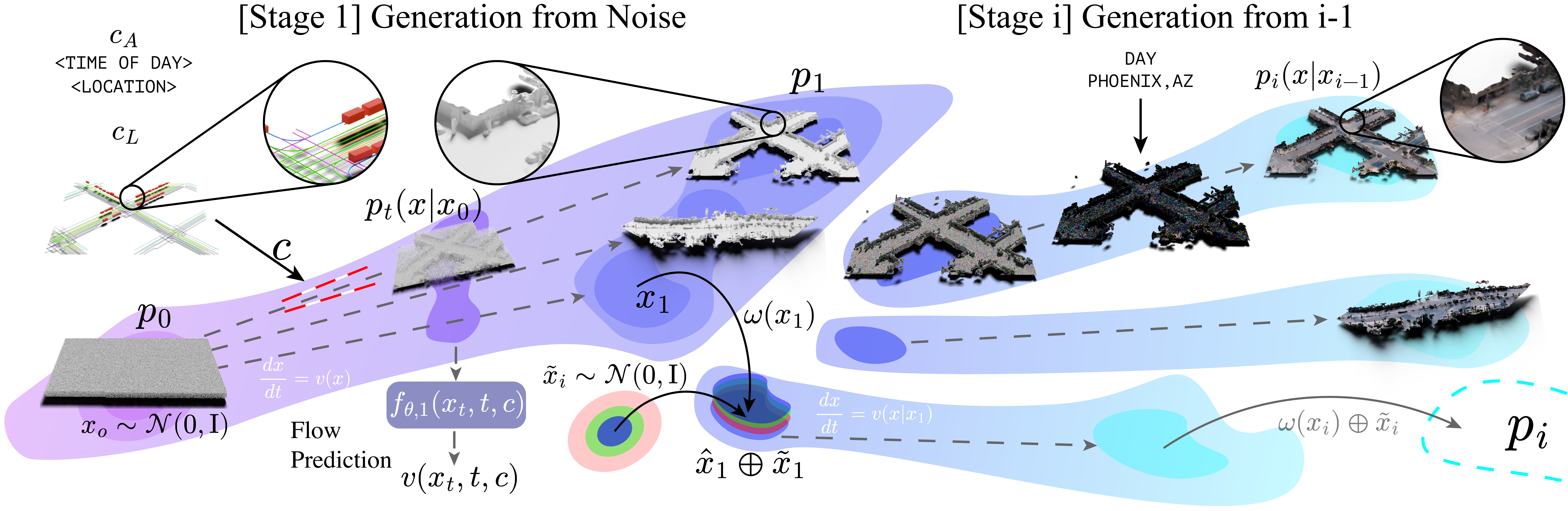}
    \caption{\textbf{WorldFlow3D} decomposes generation into a sequence of independent flows over progressively richer representations --- transporting from noise, through coarse geometry into fine geometry, and visual appearance~(\cref{ssec:flows}) . All flows operate directly in raw volumetric space (\cref{ssec:gen}), enabling a latent-free, hierarchical scene generation procedure. Generation is controlled by a vectorized geometric layout and discrete scene attributes, giving consistent structural and semantic control at every level~(\cref{ssec:control}).}
    \label{fig:method}
\end{figure*}
\paragraph{Preliminaries.}
Continuous normalizing flows (CNFs)~\cite{chen2018neural} were originally introduced to train ordinary differential equations with black-box solvers, modeling their underlying vector field $v_t$ end-to-end with deep neural networks. 
CNFs model the continuous-time flow $\phi_t$ over $t = \{0...T\}$ between two $d$-dimensional distributions $p_0$ and $p_1$ using a deep neural network $f_{\theta} (x_t, t)$ with trainable parameters $\theta$ for any sample $x \in \mathbb{R}^d$.
More recently, conditional flow matching (CFM)~\cite{lipman2022flow} methods train CNFs for optimal transport between two distributions using linear solvers.
CFMs only require samples $x$ from the underlying data distributions $p_0$ and $p_1$, and are trained to regress the underlying conditional vector field $v_t$ given sample $x_1$ of fixed conditional probability paths with

\begin{equation}\label{eq:cfm_seq}
    \mathcal{L}_\text{CFM}(\theta) = \mathbb{E}_{t, q(x_0), p_t(x|x_0)} \| f_{\theta}\left(x, t\right) - v_t\left(x  | x_0\right) \|^2,
\end{equation}
extending the scope of this approach. 

While probabilistic modeling of differential equations was also proposed for diffusion models, with stochastic differential equations~\cite{song2020denoising}, CFM has been investigated as a more efficient way to model paths between a Gaussian distribution $p_0 \sim \mathcal{N}(0, I)$ and a data distribution $p_1$ such as images or 3D data. 
Note that CFM generalizes well to arbitrary, non-diffusion probability paths such as optimal transport between \textit{any} two data distributions.
We subsequently leverage this property to propose a novel formulation of hierarchical generation.

\subsection{Flowing Through Hierarchical Data Distributions}
\label{ssec:flows}




For~\method, we define hierarchical generation as a sequence of distributions over progressively richer scene representations --- concretely, from coarse geometry to fine geometry to full appearance --- where each transition between adjacent levels corresponds to an independent learned flow.
Fig.~\ref{fig:method} depicts our approach, with separate paths indicating distinct hierarchies.
We assign a data distribution $p_i$ to each such level $i\in\left[0, N\right]$, where adjacent levels differ in fidelity and attribute composition, and therefore have distinct dimensions $d_i\ge d_{i-1}$.
We train an independent velocity field $f_{\theta, i}$ at each level $i$ to model the optimal transport of a sample $\mathbf{x}$, with the following rectified flow~\cite{liu2022rectflow} objective as
\begin{equation}
\label{eq:cfm_seq}
\begin{split}
    &\mathcal{L}_\text{CFM}(\theta_{i}) =  \\ 
    &\mathbb{E}_{t\sim\mathcal{U}[i-1, i], x \sim p_{i-1},\, x_i \sim p_i}\| f_{\theta,i}(x_t, t) - (x_i - x_{i-1}) \|^2.
\end{split}
\end{equation}

As finer-level attributes may introduce additional attributes, e.g., RGB color, we accomodate them at lower level target distributions $p_{i}$, by assuming Gaussian distributions over unknown source dimensions such that 
\begin{align}
\label{eq:adddim}
\begin{split}
    x_{i-1} &= \hat{x}_{i-1} \oplus \tilde{x}_{i-1}, \\
    &\text{with } \tilde{x}_{i-1} \sim \mathcal{N}(0, I) \in \mathbb{R}^{(d_i - d_{i-1})}
    \\ &\text{ and } \hat{x}_{i-1} \sim p_{i-1}, \in \mathbb{R}^{d_{i-1}} .
\end{split}
\end{align}

\subsection{3D Scene Generation}
\label{ssec:gen}
%
We employ the hierarchical formulation introduced in~\cref{ssec:flows} for 3D scene generation, where faithful synthesis requires resolving structure simultaneously at multiple spatial scales, and both the global spatial layout and local surface detail are necessary.

\paragraph{Latent-Free 3D Scene Representation.}
While 3D data comes in many forms (meshes, point clouds, signed distance functions), volumetric representations in particular can represent fine-grained geometric structure at a discretized voxel level, making the learning of 3D distributions $p_i$ tractable without complex compression.
Motivated by their simplicity, we choose truncated unsigned distance fields (TUDFs) to represent the surface of the scene at its zero-level set.

Generating 3D scenes that are both geometrically detailed and spatially coherent requires representations that can express structure at multiple scales of resolution and richness.
Each scene sample $\mathbf{x}_i \in \mathbb{R}^{l_i \times c_i}$, is a raw volumetric tensor of shape $l_i = X \times Y \times Z$ and attribute channels $c_i$.
Coarser levels operate at higher metric size $s_i \leq s_{i-1}$ of each individual voxel, allowing us to capture broad geometric structure.
Finer levels refine detail at lower $s_i$ and may introduce additional volumetric attributes.
At each level $i$, $x_i$ is composed of a subset of volumetric attributes defined over a voxel grid at resolution $l_i$: a TUDF $\mathbf{\mathcal{D}}_i \in [0, \tau]$.
Each voxel stores the unsigned distance to the nearest surface, truncated at $\tau$; and optionally a sparse attribute volume $\mathbf{\mathcal{C}}_i$ at the surface-set defined by $D_i\left(x,y,z\right) < \tau$. $\tau$ is the same across levels, and only varies by hierarchy.

We instantiate each sample from a distribution $p_i$ directly over all volumetric scene representations, without a latent intermediate.
Our method \textit{does not} require a latent vector produced by a latent autoencoder.
We thus achieve higher training and inference efficiency, eliminating the two-stage learning approach common in latent generation.
As we find in~\ref{ssec:ablations}, velocity prediction between \textit{geometry} distributions benefits from unmediated access to geometric structure at every point along the trajectory, avoiding the representational bottleneck and reconstruction error introduced by a learned compression.
We also note that latent-space flows remain fully compatible with the framework and may be incorporated at any stage where compression is warranted; direct volumetric generation is simply the more natural choice when the data space is tractable.

\paragraph{Flow Through the 3D Scene Hierarchy.}
In our method, the hierarchical scene representation defines a structured sequence of distributions that the generative process traverses. 
At the coarsest level, $\mathbf{v}_{\theta^{(0)}}$ transports Gaussian noise samples $x_0 \sim \mathcal{N} (0, \text{I})$ to the data distribution over coarse geometry $p_{1}$.
At each subsequent level $i$, a function $\omega_{i-1}$ may --- depending on the target distribution --- (1) spatially upsample existing attributes in $l_{i-1}$ with voxel size $s_{i-1}$ to the target tensor of shape $l_i$ with voxel size $s_i$, (2) inject additive noise to prevent mode collapse in low data regimes
\begin{align}
\begin{split}
\label{eq:bridge}
    \omega_{i-1}(x_{i-1}) &= \uparrow_r \left(x_{i-1} + \varepsilon\right), \\\textrm{where}&\quad \varepsilon \sim \mathcal{N}(0, \sigma^2I), \in \mathbb{R}^{d_{i-1}}, r = \frac{s_{i-1}}{s_i},
\end{split}
\end{align}
or (3) append independent noise channels as specified in~\cref{eq:adddim} for any new attributes introduced at level $i$, that is
\begin{equation}
    x^{(d_i)}_{i-1} = \omega_{i-1}(x^{(d_{i-1})}_{i-1}) \oplus\tilde x_{i-1}.
\end{equation}
Each learned $f_{\theta,i}$ then independently models $v_{\theta,i}(x_t^{d_i}, t)$, as described in Eq.~\ref{eq:cfm_seq}.

\subsection{Unbounded World Synthesis with Chunk-Aware Velocity Averaging}\label{ssec:unbounded}
\begin{figure}[t]
    \centering
    \includegraphics[width=\linewidth]{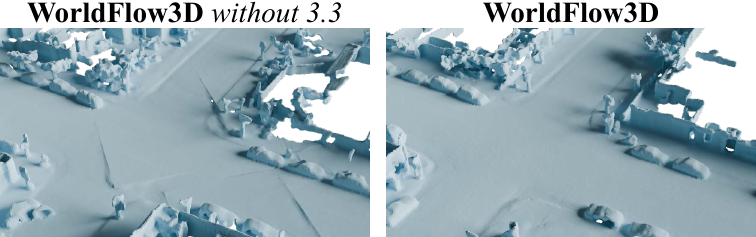}
    \caption{\textbf{Feather weighted velocity averaging} in overlapping chunk regions significantly improves the generated geometry for unbounded generations, as shown above.}
    \label{fig:weighting}
\end{figure}

Unbounded --- or even large-scale in general --- 3D scene generation is not obtainable in a single inference pass, due to compute limits set by available technology. As a result, we partition scenes into overlapping chunks $\{ \Omega_k \}$, and generate individual $\Omega_k$ in each inference pass.
Na\"ive sequential outpainting and masking~\cite{muller2024multidiff} with these chunks, however, results in artfiacts (see \cref{fig:weighting}). 
We therefore generate the full volume $x_i$ by integrating all chunks through the flow matching ODE at the same time.
At each timestep $t$, the local volume $x|_{\Omega_k}$ is extracted for all chunks and passed through the flow model $f_\theta$ to obtain the per-chunk velocity $v(x|_{\Omega_k}, t, c_k)$, where $c_k$ represents the local layout conditioning and global attributes.
All chunks at $t$ are then combined into the global velocity field $\bar{v}(t)$ via a spatially varying feather-weighted average
\begin{equation}\label{eq:vel_avg}
    \bar{v}(\mathbf{s}, t) = \frac{\sum_{k:\,\mathbf{s}\in\Omega_k} \gamma_k(\mathbf{s})\;f_\theta \left(\mathbf{x}\big|_{\Omega_k}(t),\; t,\; c_k\right)[\mathbf{s}]}{\sum_{k:\,\mathbf{s}\in\Omega_i} \gamma_k(\mathbf{s})}.
\end{equation}
\noindent Here, $\gamma_k (s)$ represents the feather weight at global location $\mathbf{s} \in \mathbb{R}^l$, which ramps linearly from a small value at chunk borders to $1$ at the center of each chunk.
This smoothly blends adjacent chunks and reduces to simple single-chunk prediction in non-overlapping regions.
The full volume is then advanced with a standard Euler integration step $ x_{t+1} = x_t - \Delta t\;\bar{v}$. 
In practice, we keep global conditioning, samples $x_t$, and computed local velocity fields $v(x|_{\Omega_k}, t, c_k)$ in CPU memory and only transfer to GPU memory for the forward pass of $f_\theta$, enabling the generation of theoretically infinite scenes, constrained only by compute resources.

\subsection{Controllable 3D Generation}\label{ssec:control}
Our method allows for explicit control over geometric structure and visual texture, a crucial requirement for usability of generated scenes.
We provide control through a geometric layout $c_L$ represented as a vectorized primitive --- polylines defining structural boundaries and bounding boxes defining object extents --- and discrete scene attributes $c_A$ encoding scene-level visual descriptors such as environment type and lighting conditions, see Fig.~\ref{fig:method}.
Typical forms of $c_L$ are room layouts or maps, while $c_A$ spans from natural text to discrete categories as presented in this work.

\paragraph{3D World Control.}
Maintaining $c_L$ in vectorized form makes it resolution-agnostic and generalizable across map formats. At each level $i$, it is voxelized on-the-fly into $c_{L,i} \in \mathbb{R}^{s_i \times K}$, where each of the $K$ channels encodes a distinct semantic class of the boundary or object type.
This allows for a single layout specification to condition generation consistently across all levels without reprocessing, and decouples the control representation from the spatial resolution of the generator.
Scene attributes $c_A \in \mathcal{A}$ are encoded as a compact embedding $c_{A, i}$ and injected globally.
In practice, we use discrete environment tags; however, $c_A$ may encode any arbitrary scene-level descriptor.


\subsection{Scale-Space Embeddings and Losses}
Each velocity field $v_i(x | c_L, c_A, x_{i-1})$ is represented by $f_{\theta,i} (x_t,t,c_L, c_A)$ as a 3D UNet with residual blocks and self-attention at multiple spatial scales.
The intermediate sample $x_t$ is concatenated channel-wise with the layout volume $c_L^{(i)}$, and the scene attributes $c^{(i)}_A$ broadcast spatially to $l_i \times \mathcal{A}$, providing the model with direct spatial access to prior-level structure, layout, and scene-level descriptors.
Each residual block applies FiLM conditioning~\cite{perez2018film} via a conditioning embedding $e^{(i)}(t)$ formed by summing independent learned embeddings of the timestep $t$, and a global layout summary $\phi_L(c^{(i)}_L)$ as
\begin{equation}
    e^{(i)}(t) = \phi_t(t) + \phi_L(c^{(i)}_L),
\end{equation}
where $\phi_t$ and $\phi_L$ are small learned encoders used to predict the scale and shift parameters of each residual block.


\subsection{Implementation Details}
The coarsest model in each sequential flow hierarchy, translating from $p^{(0)}:=\mathcal{N}(0, 1)\longmapsto p^{(1)}$, is trained for up to 1 day across 2 NVIDIA H100 GPUs (empirically, we observe saturation in quality at this point), while all subsequent flow models $p^{i-1}\longmapsto p^{i}$ are trained for 12 hours on the same infrastructure. We use the AdamW optimizer with a learning rate of $2\times10^{-6}$. We provide additional details on flow sequence modeling in the Supplementary Material.

\begin{figure*}[t]
    \centering
    \includegraphics[width=1.0\linewidth]{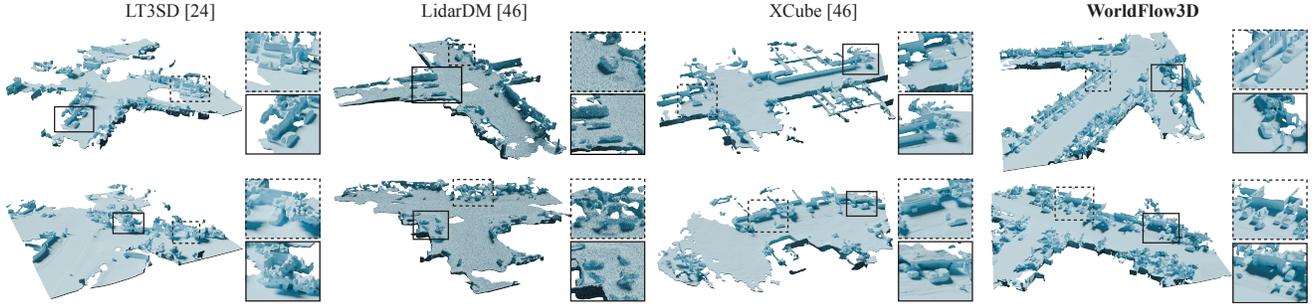}
    \caption{\textbf{Qualitative comparison on outdoor scene generation} with~\method~and baseline methods trained on the Waymo~\cite{sun2020wod} dataset. We showcase scenes generated at moderate scales, and closer-up views of specific details including buildings and vehicles. We obtain high-quality, realistic geometry and smooth surfaces with a good amount of detail, as viewed from coherent building structure, smooth road surfaces, and distinct vehicle geometry.}
    \label{fig:qual_waymo_eval}
    \vspace{-4mm}
\end{figure*}
\section{Assessment}
We evaluate the effectiveness of our method via comparisons to existing generative methods on both indoor rooms and outdoor driving environments, and both real and synthetic data distributions. 

\paragraph{Datasets.} 
We evaluate on three data distributions, using the Waymo Open Dataset~\cite{sun2020wod} as our data distribution for open-world 3D driving scenes. and the 3D-FRONT dataset~\cite{3dfront} for synthetic, indoor worlds. For all scenes, we construct volumetric TUDFs and sparse color volumes which are used as training data. For outdoor scenes, we use a hierarchy with $\{s_1, s_2\} = {0.4m, 0.2m}$ and $\tau=1m$, and for indoor scenes, we use $\{s_1, s_2\} = 0.044m, 0.022m$ and $\tau=0.1m$. We provide further detail on our data processing and parameter choices in the Supplementary Material.

\paragraph{Baselines.}
We compare against five recent methods across our target domains. For outdoor generation on Waymo, we evaluate against (a) XCube~\cite{ren2024xcube}, a hierarchical voxel latent diffusion model which set a benchmark on 3D quality, and (b) LidarDM~\cite{zyrianov2025lidardm}, using the intermediate 3D scene generation branch. For indoor generation on 3D-Front, we compare against (c) BlockFusion~\cite{wu2024blockfusion}, which generates scenes via latent triplane-based spatial outpainting, and (d) WorldGrow~\cite{worldgrow2025}, a sequential framework for unbounded 3D indoor scene synthesis. For all datasets, we compare to (e) LT3SD~\cite{meng2024lt3sdlatenttrees3d}, a latent tree-structured patch diffusion model which we re-train on Waymo and 3D-FRONT. We use official checkpoints for all other baselines, and omit InfiniCube~\cite{lu2024infinicube} and XScene~\cite{yang2025xscene} due to unavailable implementation and task mismatch respectively. Further details, including our evaluation procedure and inference procedure for baselines, are provided in the Supplementary Material.

\begin{figure}[t]
    \centering
    \includegraphics[width=1.0\linewidth]{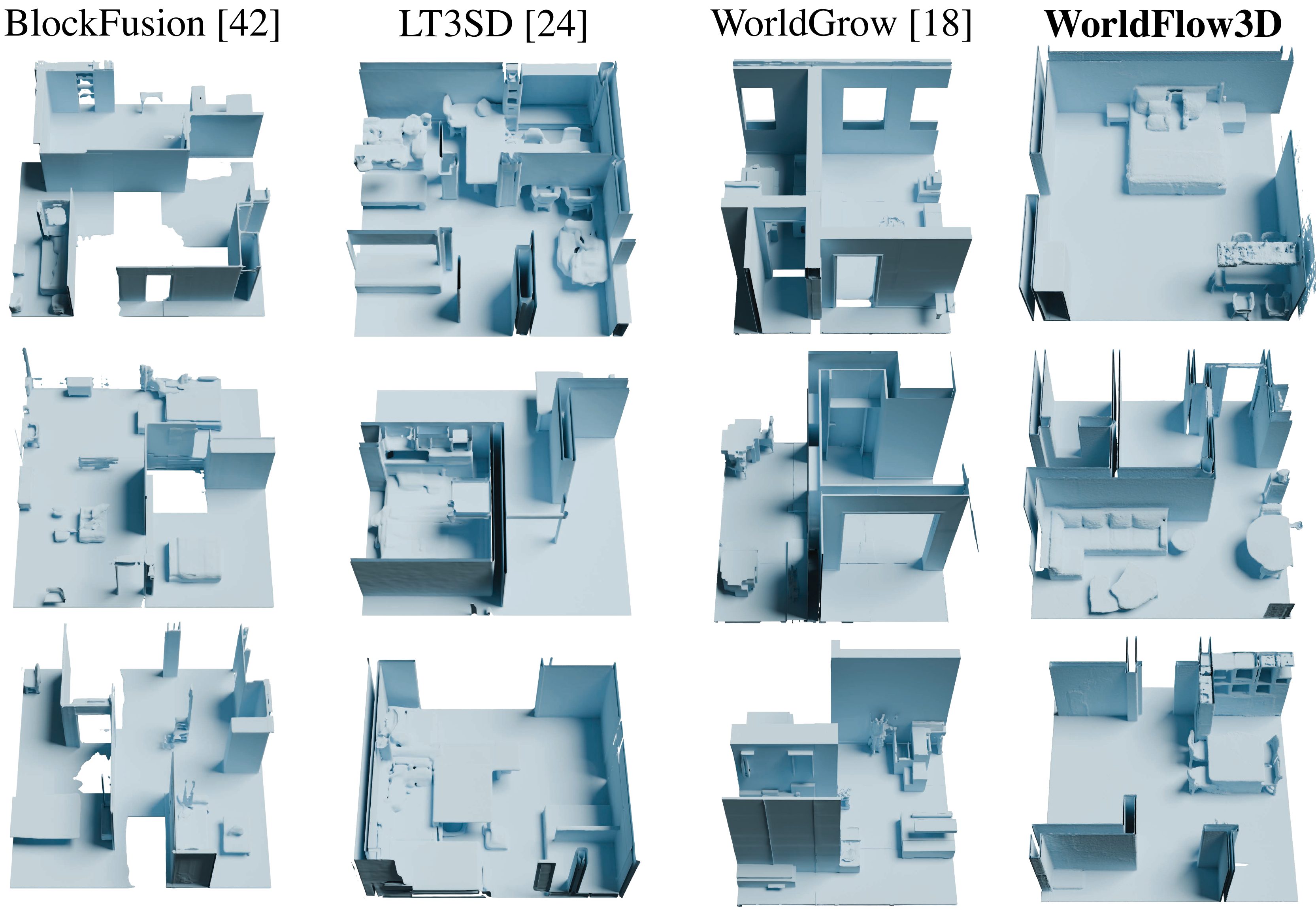}
    \caption{\textbf{Qualitative comparison on indoor scene generation} with~\method~and baseline methods trained on the 3D-FRONT~\cite{3dfront} dataset. We showcase generations of regions including (potentially multiple) rooms with various objects. Our generations are high-fidelity and contain smooth surfaces and realistic geometry.}
    \label{fig:qual_front3d_eval}
\end{figure}
\begin{table*}[t]
\renewcommand{\arraystretch}{1.2}
\centering
\small
\caption{\textbf{Quantitative Evaluation for Outdoor 3D Scene Generation} on the Waymo Open Dataset~\cite{sun2020wod} for~\method~and existing approaches. We show results for unconditional generation in the first section and conditional in the second. The best results for each metric are in \textbf{bold}; second-best are \underline{underlined}. We evaluate distribution coverage and alignment (COV, JSD), generation fidelity (MMD and 1-NNA), and feature-based distributional distance ($\mathrm{FD}_{\textrm{C}}$). We report metrics on \textit{large-scale} scene sizes of $96m\times96m$.}
\setlength\tabcolsep{3pt}
\begin{tabular*}{\linewidth}{@{\extracolsep{\fill}} l l cc cc cc c c}
    \toprule
    & \multirow{2}{*}{\textbf{Method}}
        & \multicolumn{2}{c}{\textbf{COV} $\uparrow$}
        & \multicolumn{2}{c}{\textbf{MMD} $\downarrow$}
        & \multicolumn{2}{c}{\textbf{1-NNA} $\downarrow$}
        & \multirow{2}{*}{\textbf{JSD} $\downarrow$}
        & \multirow{2}{*}{$\mathrm{FD}_{\textrm{C}}$ $\downarrow$} \\
    \cmidrule(lr){3-4}\cmidrule(lr){5-6}\cmidrule(lr){7-8}
    & & CD & EMD & CD & EMD & CD & EMD & & \\
    \midrule
    \multirow{4}{*}{\shortstack[l]{\textbf{Waymo}\\ Uncond.}}
        & XCube~\cite{ren2024xcube}
            & \underline{27.50} & \underline{24.90} & \underline{19.75} & \underline{2.99} & \underline{95.85} & \underline{85.35} & \textbf{0.480} & 214.08 \\
        & LidarDM~\cite{zyrianov2025lidardm}
            & 15.30 & 15.40 & 29.74 & 3.60 & 99.40 & 98.10 & 0.564 & 232.49 \\
        & LT3SD~\cite{meng2024lt3sdlatenttrees3d}
            & 20.00 & 16.10 & 34.33 & 3.96 & 95.85 & 85.35 & 0.524 & \underline{76.04} \\
        & \textbf{\method}
            & \textbf{33.00} & \textbf{32.70} & \textbf{16.57} & \textbf{2.81} & \textbf{89.15} & \textbf{70.15} & \underline{0.490} & \textbf{74.83} \\
    \midrule
    \multirow{2}{*}{\shortstack[l]{\textbf{Waymo}\\ Cond.}}
        & LidarDM~\cite{zyrianov2025lidardm}
            & \underline{12.00} & \underline{11.30} & \underline{35.46} & \underline{3.81} & \underline{99.60} & \underline{99.20} & \underline{0.590} & \underline{215.49} \\
        & \textbf{\method}
            & \textbf{39.70} & \textbf{35.20} & \textbf{12.44} & \textbf{2.70} & \textbf{88.60} & \textbf{81.85} & \textbf{0.483} & \textbf{80.08} \\
    \bottomrule
\end{tabular*}
\label{tab:quantitative_outdoor}
\end{table*}
\begin{table*}[t]
\renewcommand{\arraystretch}{1.2}
\centering
\small
\caption{\textbf{Quantitative Evaluation for Indoor 3D Scene Generation} on the synthetic 3D-FRONT~\cite{3dfront} dataset. The best results for each metric are in \textbf{bold}; second-best are \underline{underlined}. We report metrics on \textit{small-scale scenes}~\cite{meng2024lt3sdlatenttrees3d, wu2024blockfusion} of $2m\times 2m$.}
\setlength\tabcolsep{3pt}
\begin{tabular*}{\linewidth}{@{\extracolsep{\fill}} l l cc cc cc c c}
    \toprule
    & \multirow{2}{*}{\textbf{Method}}
        & \multicolumn{2}{c}{\textbf{COV} $\uparrow$}
        & \multicolumn{2}{c}{\textbf{MMD} $\downarrow$}
        & \multicolumn{2}{c}{\textbf{1-NNA} $\downarrow$}
        & \multirow{2}{*}{\textbf{JSD} $\downarrow$}
        & \multirow{2}{*}{$\mathrm{FD}_{\textrm{C}}$ $\downarrow$} \\
    \cmidrule(lr){3-4}\cmidrule(lr){5-6}\cmidrule(lr){7-8}
    & & CD & EMD & CD & EMD & CD & EMD & & \\
    \midrule
    \multirow{4}{*}{\shortstack[l]{\textbf{3D-Front}\\{\scriptsize\cite{3dfront}} Uncond.}}
        & BlockFusion~\cite{wu2024blockfusion}
            & 29.00 & 28.40 & 0.054 & \underline{0.221} & 91.50 & \underline{90.65} & 0.380 & 165.46 \\
        & LT3SD~\cite{meng2024lt3sdlatenttrees3d}
            & 23.40 & 26.20 & 0.056 & 0.223 & 93.00 & 91.60 & \underline{0.230} & \underline{44.97} \\
        & WorldGrow~\cite{worldgrow2025}
            & \underline{34.70} & \underline{32.80} & \underline{0.053} & \underline{0.221} & \underline{88.05} & 88.65 & 0.272 & 172.58 \\
        & \textbf{\method}
            & \textbf{38.30} & \textbf{38.10} & \textbf{0.039} & \textbf{0.195} & \textbf{74.75} & \textbf{75.80} & \textbf{0.164} & \textbf{36.45} \\
    \bottomrule
\end{tabular*}
\label{tab:quantitative_indoor}
\end{table*}
\paragraph{Evaluation Metrics.}
We evaluate generation quality using five complementary metrics: Coverage (\textsc{COV}), Minimum Matching Distance (\textsc{MMD}), 1-Nearest Neighbour Accuracy (\textsc{1-NNA}), Jensen-Shannon Divergence (\textsc{JSD}), and Fr\'{e}chet Distance ($\mathrm{FD}_{\mathrm{Concerto}}$).
\textsc{COV} measures diversity as the fraction of reference scenes $r \in \mathcal{R}$ matched by at least one generated sample; \textsc{MMD} measures fidelity as $\frac{1}{|\mathcal{R}|}\sum_{r \in \mathcal{R}} \min_{g \in \mathcal{G}} d(r, g)$; and \textsc{1-NNA} is a leave-one-out classifier over $\mathcal{G} \cup \mathcal{R}$, where an accuracy of $50\%$ indicates statistically indistinguishable distributions.
\textsc{COV}, \textsc{MMD}, and \textsc{1-NNA} are each computed under both Chamfer Distance (\textsc{CD}) and Earth Mover's Distance (\textsc{EMD}) as the underlying similarity measure $d(\cdot, \cdot)$.
\textsc{JSD} measures spatial overlap by voxelizing both sets into a shared occupancy grid and computing $\frac{1}{2}\mathrm{KL}(P \| M) + \frac{1}{2}\mathrm{KL}(Q \| M)$, where $M = \frac{1}{2}(P + Q)$.
$\mathrm{FD}_{\mathrm{Concerto}}$ computes the Fr\'{e}chet Distance between per-scene embeddings extracted from Concerto~\cite{zhang2025concerto}, a large-scale 3D foundation model pretrained on real-world point clouds, capturing high-level semantic and structural similarity beyond what point-distance metrics can express.
All metrics are computed over $N = 5{,}000$ surface points sampled per scene chunk, and we select $1,000$ scene chunks per method. Extended evaluation information and per-dataset sampling details are provided in the Supplementary Material.

\subsection{3D Generation Results}
We conduct a quantitative evaluation of the 3D generation quality of our method and competing baseline approaches. In Tab. ~\ref{tab:quantitative_outdoor}, we provide results for outdoor scene generation on the Waymo dataset, both unconditionally and conditionally. In Tab. ~\ref{tab:quantitative_indoor}, we compare our method for indoor scene generation to baseline results on the 3D-Front dataset. Across multiple data distributions,~\method~ outperforms existing baselines in all quantitative evaluations, demonstrating not only high geometric \textit{fidelity} but also a high degree of geometric \textit{diversity}. We demonstrate examples of very large-scale generated scenes in Fig.~\ref{fig:teaser} to supplement these numerical results, exhibiting the core elements of our method: large-scale, effectively unbounded scenes, explicit scene control, high-fidelity scene geometry, and visual attributes such as texture. 
In Figures~\ref{fig:qual_waymo_eval} and~\ref{fig:qual_front3d_eval}, we ground this with further visual comparisons to the existing baselines we quantitatively compared to, supplementing our demonstration of superior quality. While competing methods demonstrate reasonable fidelity, ours attain higher levels of quality and 3D consistency. Our training objective results in broader generalizability and this is reflected in higher distribution coverage across all datasets. 
Furthermore, we do not compare relative to our discretized training distribution as in existing methods~\cite{meng2024lt3sdlatenttrees3d, worldgrow2025}, which is inherently lower resolution -- but \textit{we compare to the original, arbitrarily higher-resolution data} as our ground truth. This inherent difficulty is especially present for outdoor scene generation, as shown in the difference in evaluation metrics compared to indoor data. Nevertheless, our approach obtains reasonable distribution coverage and reasonable feature-space similarity; furthermore, visual evaluations (see~\ref{ssec:user_study}) confirm the quality of our generations in their geometric structure.

\begin{table*}[t]
\renewcommand{\arraystretch}{1.2}
\centering
\small
\caption{\textbf{Ablation Study} over the core contributions of our method, comparing distribution coverage (COV), geometric fidelity (MMD and 1-NNA), visual texture quality ($\mathrm{FD}_{\textrm{C}})$, and training efficiency (GPU-hrs). We compare against traditional latent diffusion and latent flow approaches, followed by an ablation of our flow matching through distributions objective, and finally of our flow sequence hierarchy. We conduct our evaluation on the Waymo Open Dataset~\cite{sun2020wod}, and the best results are \textbf{bolded}.}
\setlength\tabcolsep{3pt}
\begin{tabular*}{\linewidth}{@{\extracolsep{\fill}} l cc cc cc c c}
    \toprule
    \multirow{2}{*}{\textbf{Ablation}}
        & \multicolumn{2}{c}{\textbf{COV} $\uparrow$}
        & \multicolumn{2}{c}{\textbf{MMD} $\downarrow$}
        & \multicolumn{2}{c}{\textbf{1-NNA} $\downarrow$}
        & \multirow{2}{*}{$\mathrm{FD}_{\textrm{C}}$ $\downarrow$}
        & \multirow{2}{*}{\textbf{GPU-hrs} $\downarrow$} \\
    \cmidrule(lr){2-3}\cmidrule(lr){4-5}\cmidrule(lr){6-7}
    & CD & EMD & CD & EMD & CD & EMD & & \\
    \midrule
    Latent Diffusion                      & 37.75 & \underline{33.75} & 7.426 & 1.801 & 91.38 & 75.63 & 120.87 & 288 \\
    Latent Flow                           & 37.00 & 32.75 & \underline{6.755} & 1.778 & \underline{90.88} & \underline{74.88} & 112.35 & 288 \\
    Flow from Noise                       & 33.75 & \underline{33.75} & 6.887 & \underline{1.747} & 92.25 & 81.25 & 107.59 & \underline{144} \\
    WorldFlow3D ($p^{(f)}=p^{(1)}$)       & \underline{38.00} & 31.50 & 7.101 & 1.821 & 92.75 & 75.63 & \textbf{89.47} & \textbf{72} \\
    \textbf{\method (full)}               & \textbf{42.50} & \textbf{37.25} & \textbf{4.942} & \textbf{1.696} & \textbf{85.00} & \textbf{70.25} & \underline{103.20} & \underline{144} \\
    \bottomrule
\end{tabular*}
\label{tab:ablation}
\end{table*}
\begin{figure*}[t!]
    \centering
    \includegraphics[width=1\linewidth]{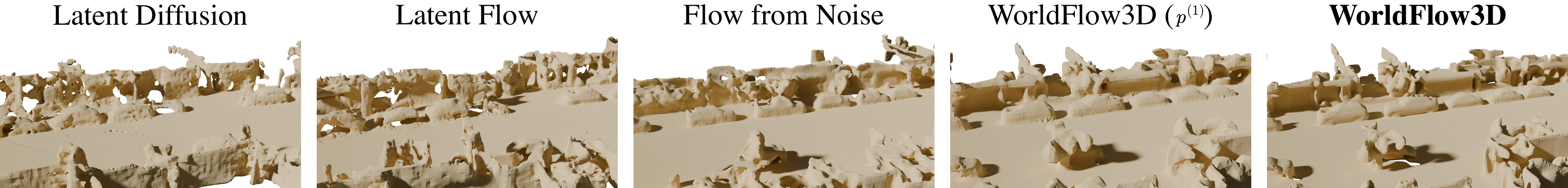}
    \caption{\textbf{Ablation of Core Components.} We provide qualitative results obtained by ablating the key components of our method. Latent diffusion and flow approaches produce structurally degenerate, non-realistic results, while flow from noise at finer distributions produces noisy outputs. Examples of this can be seen on the building walls, which are smooth with~\method, and on vehicle details such as tires. Our hierarchical, latent-free approach is the only one that can produce high-quality, geometrically plausible results.}
    \label{fig:ablations}
    \vspace{-5mm}
\end{figure*}

\begin{figure*}[t]
    \centering
    \includegraphics[width=\linewidth]{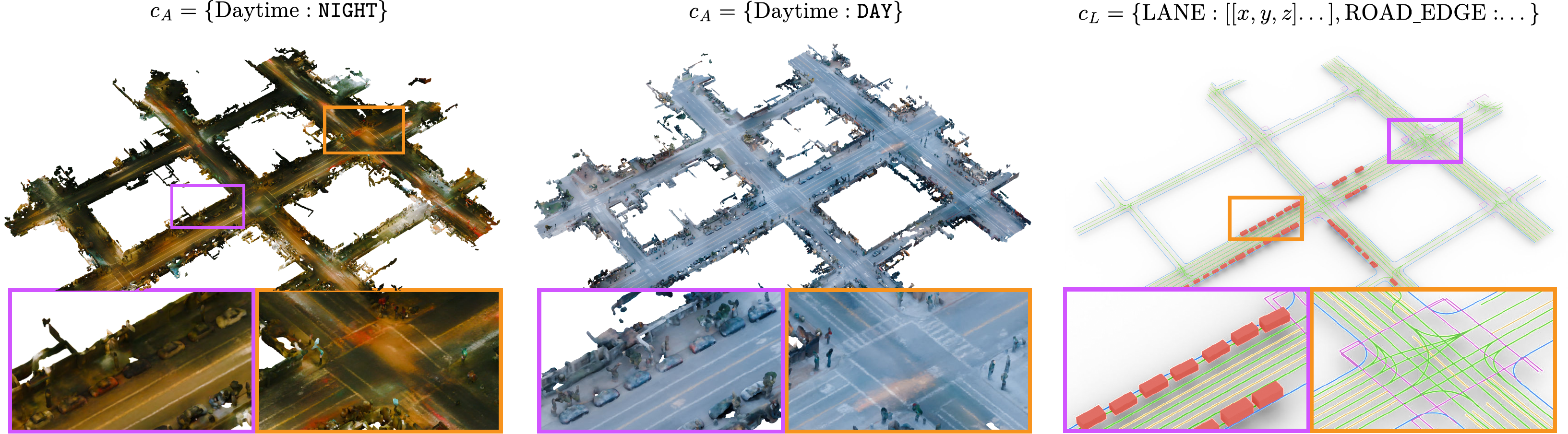}
    \caption{\textbf{Visual Texture and Controllability.} We report large-scale outdoor scenes with texture control via text attributes and geometry control via vector maps, yielding results conditioned on the same geometry to produce distinct environments.}
    \label{fig:texture_control}
    \vspace{-4mm}
\end{figure*}

\paragraph{Controllability Evaluation.}
We provide additional qualitative results which validate our method's controllability in Figure~\ref{fig:texture_control} --  confirming, respectively, attribute control over visual texture and fine-grained geometric structure control using road map layouts. Our generated scenes not only strictly adhere to control, on the level of individual objects (such as vehicles, for road layouts), but are visually expressive for distinct textures, showcasing diversity.

\subsection{Ablation Study}\label{ssec:ablations}
On a foundational level, we compare standard latent diffusion or latent flow approaches to our proposed latent-free generative method. We conduct our ablation study on the Waymo dataset. 
We compare our method to traditional
latent-space generation approaches with \texttt{Latent Diffusion} and \texttt{Latent Flow} --- using the same hierarchical structure as our main results, but incorporating a VAE at both levels and performing diffusion or flow, respectively, in latent space. 
We compare additionally the results of flowing from noise (conditional denoising) as opposed to flowing through distributions in \texttt{Flow From Noise}.
Finally, we ablate our hierarchical structure by comparing the results of~\method~after only one distribution.
The results in Figure~\ref{fig:ablations} visually demonstrate that our latent-free flow through distributions is most capable of producing realistic, high-quality geometry, and a quantitative evaluation in Tab. ~\ref{tab:ablation} confirms this. We conduct this evaluation using smaller scene sizes than the results in Tab. ~\ref{tab:quantitative_outdoor}, hence the distinct value range. This is in order to focus metric variation more specifically on geometric quality, and this is evidenced by metric variation in 1-NNA and MMD. Our approach outperforms latent-based methods and is dramatically more efficient, as we discuss in the following section.


\paragraph{Flowing Through Distributions vs. Conditional Denoising.}
The value of our approach of flowing through distributions as opposed to the traditional formulation of successive conditional denoising is most strongly confirmed in the comparison shown in Figure~\ref{fig:ablations}. Reducing the transport between distributions by flowing from an intermediate $p^i$ rather than conditionally flowing from new noise allows the model to focus on generating detail rather than structure, such as the tires on the vehicles. This is also evidenced in visual color quality (see the Supplementary Material), further confirming the usefulness of our novel approach.

\subsection{Training Efficiency}
\method~is at least 2$\times$ more efficient in training compared to traditional generative approaches, as a result of our latent-free paradigm and approach which minimizes transport between finer distributions. In contrast to existing methods which require multiple days of sequential autoencoder and latent generative model training~\cite{meng2024lt3sdlatenttrees3d, ren2024xcube, worldgrow2025}, our method requires no autoencoder and can converge on complex data distributions rapidly, reaching high levels of fidelity within only a couple of hours for finer distributions $p^{i}$ and requiring less than a day for full convergence (as validated quantitatively). In fact, our flow models between finer distributions converge to high quality in only $12$ hours of training. In contrast, as in Tab. ~\ref{tab:ablation}, even our comparatively efficient latent generation approaches (as conducted for our ablation) require two-stage training which takes $2\times$ as long as our own two-level hierarchical flow. Existing baselines, furthermore, require multiple days, coming to over a week for certain methods~\cite{meng2024lt3sdlatenttrees3d}. As a result, our method is not only capable of producing higher-fidelity results, but also accomplishes this with much lower computational cost for model training.

\begin{table}[t]
\centering
\small
\caption{\textbf{User Study.} We report Bradley-Terry (BT)~\cite{bradley1952rank} scores with 95\% bootstrap confidence intervals and overall win rates.
  }
\centering
\resizebox{\columnwidth}{!}{
    \begin{tabular}{lcccc}
      \toprule
      \textbf{Method} & \textbf{Rank} & \textbf{BT Score} & \textbf{95\% CI} & \textbf{Win Rate} \\
      \midrule
      \textbf{WorldFlow3D} &  \textbf{1} &  \textbf{0.692} &  \textbf{[0.569, 0.813]} &  \textbf{88\%} \\
      XCube~\cite{ren2024xcube}     & 2 & 0.212 & [0.118, 0.314] & 63\% \\
      LT3SD~\cite{meng2024lt3sdlatenttrees3d}     & 3 & 0.085 & [0.047, 0.131] & 43\% \\
      LidarDM~\cite{zyrianov2025lidardm}   & 4 & 0.011 & [0.004, 0.022] &  6\% \\
      \bottomrule
    \end{tabular}
}
\label{tab:user_study}
\end{table}

\subsection{User Study}\label{ssec:user_study}
We further supplement our evaluation by conducting a two-alternative forced choice user study for outdoor scene. 
Participants compared pairs of extracted meshes generated by all methods in~\cref{tab:quantitative_outdoor} and selected the one which they perceived as higher quality. 
We then fit a Bradley-Terry model~\cite{bradley1952rank} to the comparison data, and obtain a global quality score for each method in~\cref{tab:user_study}, estimating confidence intervals via bootstrapping with $n=1000$. Additional pairwise win rates with binomial significance tests and a detailed study setup are provided in the Supplementary Material.
Overall, users prefer the results of our method with high significance over all other baseline methods, providing further perceptual confirmation of our  qualitative (\cref{fig:qual_waymo_eval}) and quantitative (\cref{tab:quantitative_outdoor}) results.
\section{Conclusion}
In this work, we revisit 3D generation and model it more generally as a problem of flowing through hierarchical distributions. In this paradigm, we present~\method, a novel approach capable of producing unbounded 3D worlds with explicit scene control and high-quality geometry and texture. We validate~\method~across distinct data distributions, including \textit{both} real and synthetic data, confirming our method's generalizability, fidelity, and efficiency. The generality of our flow through distributions approach opens the door to future work using flow matching to transport between even more complex distributions and scene representations, including animated 3D scenes and radiance fields, and, as such, we believe~\method~is an innovative step towards 3D world generation.

\section*{Acknowledgements}

Felix Heide was supported by an NSF CAREER Award (2047359), a Packard Foundation Fellowship, a Sloan Research Fellowship, a Sony Young Faculty Award, a Project X Innovation Award, a Amazon Science Research Award, and a Bosch Research Award. Felix Heide is a co-founder of Algolux (now Torc Robotics), Head of AI at Torc Robotics, and a co-founder of Cephia AI.

%
%
\bibliographystyle{splncs04}
\bibliography{main}
\end{document}